\pdfoutput=1

\documentclass[11pt]{article}

\usepackage{acl}

\usepackage{times}
\usepackage{latexsym}

\usepackage[T1]{fontenc}

\usepackage[utf8]{inputenc}

\usepackage{microtype}

\usepackage{inconsolata}

\usepackage{graphicx}
\usepackage{algorithm} 
\usepackage{algorithmic}
\usepackage{arydshln}
\usepackage{comment}
\usepackage{multirow}
\usepackage{booktabs}
\usepackage{subfigure}
\usepackage{amsmath}

\usepackage{xcolor}

\usepackage{hyperref}
\hypersetup{
    colorlinks=true,
    linkcolor=blue,
    urlcolor=magenta,
    filecolor=magenta,      
    pdftitle={Overleaf Example},
    pdfpagemode=FullScreen,
    }

\definecolor{danshuhong}{RGB}{234, 107, 102}

\title{Which Side Are You On? A Multi-task Dataset for End-to-End Argument Summarisation and Evaluation}

 \author{Hao Li$^\clubsuit$\footnotemark[1], Yuping Wu$^\clubsuit$, Viktor Schlegel$^{\diamondsuit \clubsuit}$,
 \\ {\bf Riza Batista-Navarro$^\clubsuit$}, {\bf Tharindu Madusanka$^\clubsuit$}, {\bf Iqra Zahid$^\clubsuit$}, {\bf Jiayan Zeng$^\spadesuit$}
 \\ {\bf Xiaochi Wang$^\spadesuit$}, {\bf Xinran He$^\spadesuit$}, {\bf Yizhi Li$^\clubsuit$} \and {\bf Goran Nenadic$^{\clubsuit}$} \\
         $^\clubsuit$ The University of Manchester, United Kingdom  \\
         $^\diamondsuit$ASUS Intelligent Cloud Services (AICS), Singapore\\
         $^\spadesuit$The University of Leeds, United Kingdom \\
         }

\begin{document}

\maketitle
\renewcommand{\thefootnote}{\fnsymbol{footnote}} 
\footnotetext[1]{Corresponding email: \texttt{hao.li-2@manchester.ac.uk}}
\begin{abstract}
With the recent advances of large language models (LLMs), it is no longer infeasible to build an automated debate system that helps people to synthesise persuasive arguments. Previous work attempted this task by integrating multiple components. In our work, we introduce an argument mining dataset that captures the end-to-end process of preparing an argumentative essay for a debate, which covers the tasks of \textbf{claim} and \textbf{evidence identification} (Task 1 ED), \textbf{evidence convincingness ranking} (Task 2 ECR), \textbf{argumentative essay summarisation} and human preference ranking (Task 3 ASR) and \textbf{metric learning} for automated evaluation of resulting essays, based on human feedback along argument quality dimensions (Task 4 SQE).  Our dataset contains 14k examples of claims that are fully annotated with the various properties supporting the aforementioned tasks. We evaluate multiple generative baselines for each of these tasks, including representative LLMs. We find, that while they show promising results on individual tasks in our benchmark, their end-to-end performance on all four tasks in succession deteriorates significantly, both in automated measures as well as in human-centred evaluation. This challenge presented by our proposed dataset motivates future research on end-to-end argument mining and summarisation. The repository of this project is available at \url{https://github.com/HaoBytes/ArgSum-Datatset}

\end{abstract}

\section{Introduction}

\begin{figure*}[t!]
    \centering
    \includegraphics[scale = 0.75]{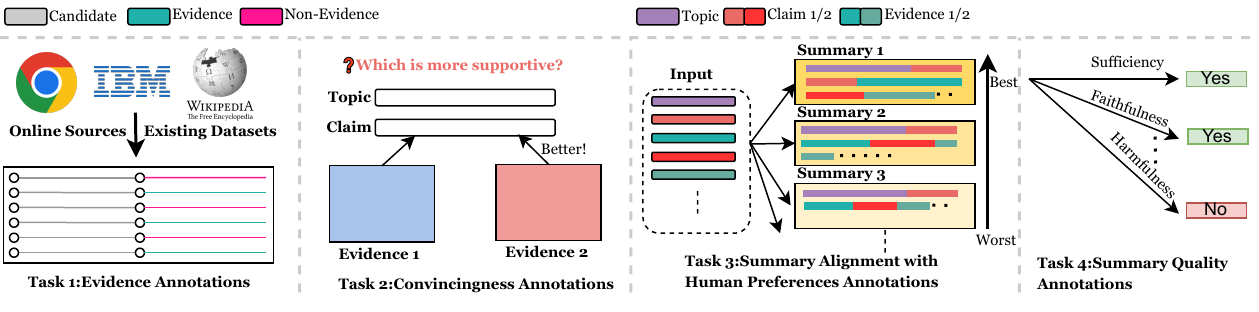}
    \caption{Overview of proposed annotation pipeline, which includes four main tasks. Task 1 identifies whether a snippet is an evidence for a given claim; Task 2 selects the appropriate evidence for each claim to make it the most persuasive; Task 3 generates a diversity of debate scripts for a given debate topic and stance then ranks them according to human preference, with the dimension of quality being measured in Task 4.
    }
    \label{overview}
\end{figure*}

Crafting arguments for a debate requires a wealth of knowledge, meticulous logical thinking and a high level of command 
of language \citep{field2017debating}. Recent trends treat this task as the automatic summarisation of a large body of debate-related material to alleviate the effort and time spent on brainstorming by stakeholders \citep{DBLP:journals/corr/abs-2011-07251}. Most of previous work split these processes into multi-component systems, including context-dependent claim detection (CDCD) \citep{DBLP:conf/emnlp/DaxenbergerEHSG17, DBLP:conf/acl/ReimersSBDSG19, DBLP:conf/acl/ChengWBS20}, claim stance classification (CSC) \citep{DBLP:conf/eacl/SlonimBSBD17, DBLP:conf/aaai/TrautmannDSSG20, DBLP:conf/emnlp/Toledo-RonenOBS20, DBLP:conf/clef/RajulaHP22}, context-dependent evidence detection (CDED) \cite{DBLP:conf/nldb/IskenderSP021,DBLP:conf/aaai/Ein-DorSDHSGAGC20, DBLP:conf/acl/ShnarchPDGHCAS18, DBLP:conf/acl/AharoniPLHLRGS14} and argument summarisation (AS) \citep{DBLP:conf/acl/Bar-HaimEFKLS20, DBLP:conf/emnlp/Bar-HaimKEFLS20, DBLP:journals/ail/YamadaTT19,DBLP:conf/sigdial/MisraEW16}. However, thus far, these tasks have been approached in isolation without formulating them into an end-to-end process. 
This contradicts the natural sequence of actions in the debate preparation process \citep{rear2017critical}, where outputs of previous tasks (e.g. finding relevant evidence) form the basis for the subsequent tasks (e.g. arranging claims and evidence in an argumentative essay).

The only comprehensive automated debating work at present is IBM's Project Debater \citep{slonim2021autonomous}. However, it is composed of multiple independent components, potentially giving rise to cascading errors, whereby errors made in any of the preceding components might be propagated and amplified in subsequent components \cite{kleinberg2007cascading}. Currently, there is no dataset supporting the development and evaluation of end-to-end systems. 
Furthermore, Project Debater lacks evaluative components; its argument quality component only considers how relevant potential candidate texts are to the debate topic. 

Earlier works explored the assessment of the argument quality in various dimensions, like convincingness of arguments and evidence \citep{DBLP:conf/chi/WambsganssNC0HL20, DBLP:conf/acl/GleizeSCDMAS19, DBLP:conf/argmining/PotashFH19, DBLP:conf/acl/HabernalG16, DBLP:journals/ress/CyraG11} and sufficiency checking \citep{DBLP:conf/argmining/GurckeAW21, wachsmuth2020intrinsic, DBLP:conf/argmining/PotashFH19, DBLP:journals/coling/StabG17, DBLP:phd/dnb/Stab17}. Recent advances in large-scale language models (LLMs) such as \texttt{GPT-3.5}\footnote{https://chat.openai.com\label{chatgpt}}, \texttt{GPT-4} \citep{DBLP:journals/corr/abs-2303-08774}, \texttt{LlaMA} \citep{DBLP:journals/corr/abs-2302-13971} and \texttt{Bard} \citep{DBLP:journals/corr/abs-2305-10403}, resulted in the achievement of seemingly human-like generated text according to evaluation dimensions such as fluency and richness \citep{zhao2023survey}. As such, the quality of their output falls short with respect to traditional argumentative quality criteria such as convincingness \citep{habernal2016argument}. At the same time, LLMs still exhibit deficiencies, such as the generation of false information \citep{azamfirei2023large, DBLP:conf/bionlp/LiWSBNKZB0N23}. \citet{DBLP:journals/corr/abs-2304-07666} proposed a GPT-generated argumentative essay corpus with annotations by human experts to understand the lexical, syntactic and stylistic features of AI-generated content (AIGC). However, they do not take into account fact-checking the generated texts and assessing the quality of essays in terms of argumentative properties. Such evaluations are crucial in argument mining, as using false evidence to convince others can lead to untenable decisions. 
 
To address these shortcomings, we have developed an end-to-end dataset that cohesively unifies these primary tasks and provides a comprehensive view of debate preparation (illustrated in Figure \ref{overview}). We provide human annotations for each of the involved tasks as well as explore the performance of SotA LLMs on these tasks in an end-to-end manner, facilitating design and evaluation of automated debate assistant systems, aligned with human preferences.
Our proposed dataset supports tasks that are broadly divided into two streams:
\emph{(1)}  evidence detection (ED), \emph{(2)}  evidence convincingness ranking (ECR) and \emph{(3A)}  argument summarisation (AS), are concerned with finding evidence to support the arguments and arranging them into a narrative. Meanwhile, \emph{(3B)} argument summarisation ranking (ASR), learning to rank generated summaries from human preferences and \emph{(4)}  learning automated metrics for summary quality evaluation (SQE), are concerned with automating fine-grained evaluation of the generations from a (weak) human supervision signal. 
In particular, we design ED, ECR and SQE as classification tasks.
AS is formulated as a summarisation task, aimed at generating debate scripts. 
We conduct an extensive empirical evaluation of the selected baselines on the proposed tasks. 
In addition, we rely on human assessments to verify the quality of the generated tasks, which showed a positive correlation between automated evaluation and human preference. In summary, the main contributions of this paper are: 
\begin{itemize}
    \item A benchmark dataset named Argument Summarisation and Evaluation (ASE) with four sub-tasks which cover the whole debate preparation process, and benchmarks for the proposed tasks with the LLMs.
    \item Investigation of the feasibility and performance of an-end-to-end debate preparation assistant, evaluated by humans and automatic metrics based on human preference. 
\end{itemize}

\section{Related Work}

One of the challenges in the field of argument mining is the lack of high-quality annotated arguments to serve as training and test data. Some work has been done to alleviate the situation, focussing on different sub-tasks such as claim identification \citep{DBLP:conf/emnlp/RinottDPKAS15, DBLP:conf/coling/LevyBGAS18, DBLP:conf/emnlp/ShnarchCMAS20}, stance identification \citep{DBLP:conf/eacl/SlonimBSBD17,DBLP:conf/lrec/JochimBBS18, DBLP:conf/coling/Toledo-RonenBHJ18}, evidence identification \citep{DBLP:conf/acl/ShnarchPDGHCAS18, DBLP:conf/aaai/Ein-DorSDHSGAGC20}, automatic identification of relational properties \citep{DBLP:conf/emnlp/HabernalG15,DBLP:conf/argmining/LawrenceR17,DBLP:conf/emnlp/LiDC20,liu2023argument,liu2023global}, argument summarisation and clustering \citep{DBLP:conf/sigdial/MisraEW16, DBLP:conf/acl/ReimersSBDSG19,DBLP:conf/emnlp/AjjourAWS19, syed-etal-2023-frame}, key point analysis \citep{DBLP:conf/acl/Bar-HaimEFKLS20,DBLP:conf/argmining/FriedmanDHAKS21}, and debate speech analysis \citep{DBLP:conf/emnlp/MirkinMOKKLJBAS18, DBLP:conf/argmining/LaveeOKKGDJBAS19, DBLP:conf/emnlp/OrbachBGKDLKMJA19, DBLP:conf/acl/OrbachBTLJAS20}. Others contributed to evaluation of argument quality, like convincingness ranking \citep{DBLP:conf/acl/GleizeSCDMAS19, DBLP:conf/emnlp/ToledoGCFVLJAS19, DBLP:journals/corr/abs-2205-07938}, sufficiency checking  \citep{DBLP:conf/argmining/GurckeAW21, wachsmuth-werner-2020-intrinsic, DBLP:conf/argmining/PotashFH19, DBLP:journals/coling/StabG17, DBLP:phd/dnb/Stab17} or fact-checking \citep{azamfirei2023large}. However, all of these work focus on only one of the above-mentioned tasks, making them difficult to integrate due to the different data sources.
Similar to our work is the IAM dataset \citep{DBLP:conf/acl/ChengBHYZS22}, but there, only arguments at the sentence-pair level are considered, building \emph{<claims, positions, evidence>} tuples, instead of providing a document-level narrative to provide a comprehensive understanding; also, the resulting argument quality is not assessed. In addition, the proposed task focusses on long articles, whereas we focus on shorter, social-media style texts with non-standard structures. 

\section{Argument Summarisation and Evaluation Dataset}

In this section, we discuss the process of data collection and annotation. The annotation process is divided into two main stages, each with two sub-tasks: Stage 1 includes the classification of evidence and the ranking of the persuasiveness of the evidence; Stage 2 is focussed on the quality of the generated summaries.

\subsection{Data Collection}
\textbf{Evidence Collection:}
We collected 1881 pieces of evidence candidates supporting claims across 31 debate topics. To align well with existing argument mining resources, all debate topics and their underlying claims and non-argumentative texts are taken from the English ArgKP dataset \citep{DBLP:conf/acl/Bar-HaimEFKLS20}. Human annotators were asked to collect text snippets that are likely to support a given claim under the debate topic. Text snippets were allowed to be reused for multiple claims and are sourced from various open-source online forums, news, as well as pre-existing datasets. In our dataset, the URL of the original source is clearly indicated for each evidence candidate.

\textbf{Summary Collection:} Each summary is a narrative consisting of \emph{(a)} all main claims (also known as key points \citep{DBLP:conf/emnlp/Bar-HaimKEFLS20}) for the same stance on the same topic taken from \citet{DBLP:conf/acl/Bar-HaimEFKLS20}'s dataset and  \emph{(b)} the evidence to support that claim. We refrain from relying on human annotators to write reference summaries directly, as this may exhibit considerable variability in the outcomes due to differences in writing style or knowledge limitations. Instead, we use LLMs (including \texttt{GPT-4} \citep{DBLP:journals/corr/abs-2303-08774}, \texttt{GPT-3.5}\textsuperscript{\ref{chatgpt}},  \texttt{GPT-3} \citep{DBLP:conf/nips/BrownMRSKDNSSAA20}, \texttt{Bard}\footnote{The backbone of the Bard at the completion of this work was PaLM2 instead of Gemini} \citep{DBLP:journals/corr/abs-2305-10403}, \texttt{LlaMA-65B} \citep{DBLP:journals/corr/abs-2302-13971}, \texttt{GLM-130B} \citep{DBLP:journals/corr/abs-2210-02414}, \texttt{Vicuna-13B}\footnote{https://lmsys.org/blog/2023-03-30-vicuna/}, \texttt{Alpaca-13B}\footnote{https://crfm.stanford.edu/2023/03/13/alpaca.html} and \texttt{BLOOM-176B} \citep{DBLP:journals/corr/abs-2211-05100}) to automatically generate the summaries, which are then evaluated by human annotators, both intrinsically (by ranking the summaries by preference) and extrinsically (by assigning scores along established argument quality dimensions). 

\subsection{Data Annotation}
\textbf{Task 1: Evidence Annotations:} Using the Amazon Mechanical Turk (AMT) crowd-sourcing platform\footnote{https://www.mturk.com/}, we collected gold labels to associate the collected evidence with the arguments. For each piece of evidence, given the debate topic and argument as context, annotators were asked to answer the following two questions: \emph{``Would you think the snippet is relevant to the argument?''} and \emph{``Would you think the snippet supports the argument?''}. They were directed to label all evidence pieces relevant to the argument, and only if the answer to both questions was ``yes'' then it was considered to be a piece of evidence supporting the argument (see example in Appendix \ref{sec:template}). Each piece of evidence was annotated by at least four annotators. The ground truth label was decided by majority vote.

\textbf{Task 2: Evidence Convincingness Annotations:} Following \citet{DBLP:conf/aaai/GretzFCTLAS20} and \citet{DBLP:conf/emnlp/ToledoGCFVLJAS19}, annotators were presented with a binary question per argument, asking \emph{``In a conversation about the topic, where you can only give a single evidence out of the following two, which one would you rather use?''}. Texts involved in the ranking are all the text snippets instead of the gold label obtained in Task 1 so that both ground truth-based and pipeline-based setting could be measured. After the annotation process was finished, we applied \texttt{milp\_ranker}\footnote{https://github.com/twoertwein/milp\_ranker} to derive a ranking of all related pieces of evidence for an argument from corresponding pairwise comparisons we get from the annotations; the implementation details can be found in Appendix~\ref{sec:appendix_milp_ranker}. To align with annotated results in Task 1, irrelevant evidence (labelled as 0 in Task 1) was removed from the final ranked lists.

\textbf{Task 3A: Argumentation Summary:}
Each topic was divided into two stances separately, to generate summaries using each of the models mentioned in Section 3.1 in different scenarios. Instead of giving definitive quantitative evidence, we instructed the model to select the appropriate evidence(s) to generate summaries, following the intuition that humans choose the most effective evidence to convince the audience \citep{lipton1993best}, during preparation, instead of relying on the quantity of evidence. However, the exact number of distinct pieces of evidence required to be convincing is an open question \citep{schwardmann2022self}. Specifically, 
we supply $n$  pieces of evidence from $S$ sources as input, where $n$ is either \emph{one}, \emph{two} or \emph{all} evidence pieces and $S$ is either the gold standard evidence matched with a claim or predicted by the best-performing model on Task 2. 
We restricted our model choice to instruction fine-tuned models, to minimise the impact of prompt engineering, as performance of non-instruction-tuning models depended heavily on the choice of prompt and in-context-learning examples (e.g., \texttt{GLM-130B}, \texttt{BLOOM-176B} repeated input and \texttt{LlaMA-65B} summarised it into a single sentence).

\textbf{Task 3B: Summary Ranking based on Human Preference:} To provide assistance in training helpful, truthful and harmless tools \citep{DBLP:journals/corr/abs-2303-08774}, such as a debate assistant, we defined the evaluation of machine-generated summaries as a ranking task aligned with human preferences. In order to provide annotators with consistent assessment criteria and quantitative performance, we ask them to consider \textsc{Helpful}, \textsc{Truthful}, \textsc{Fluent} and \textsc{Harmless} as the dimensions that measure the quality of the generated text \citet{DBLP:conf/nips/Ouyang0JAWMZASR22}. We further use \textsc{Significant}, \textsc{Informative}, \textsc{Faithfhul} and \textsc{Sufficient} as dimensions to measure the quality of the summary as an argumentative text~\citep{DBLP:conf/acl/LiSBN23}. Detailed definitions of the dimensions can be found in Appendix~\ref{sec:quality_summaries}. 

The annotators were asked to rate the outputs on a Likert scale ranging from ``The Best'' ($5$) to ``The Worst'' ($1$) for each of the scenarios, with each score to be assigned only once. 
Finally, the annotators had to rank all 36 summaries. To obtain fine-grained rankings, we form the data into the ranked instances with at least two values separated by more than 1 and a total length of 4 that are sufficiently distinguished but when assembled together could show absolute ranking (for instance, given a ranking data set \emph{A<B<C<D<E<F}, we split it into \emph{A<B<D}; \emph{B<C<E}; \emph{C<D<F}; \emph{D<E<F}). Based on the fact that each annotator's preference may differ from others as well as the diversity in the models' outputs, we normalise the performance of the model into scores using the Bradley–Terry model~\cite{guo2018experimental} and the two-parameter logistic model following the Item Response Theory (IRT) \cite{embretson2013item} separately. The former quantifies the performance of the model on specific settings by pairwise comparisons, while the latter allows for an overall measure of the level of difficulty of the different settings; the detailed explanation and ablation experiments of both models can be find in Appendix~\ref{sec:bt_irt}.

\textbf{Task 4: Summary Quality Annotations:} This task requires to rank the outputs and is simultaneously carried out with Task 3. We asked the annotators to assess factual errors, sufficiency (i.e. whether the given text can persuade readers to accept its stance) and logical coherence of each summary when ranking the summaries. They needed to identify the errors as described above and then mark their location. 


\subsection{Annotation Statistics and Analysis}
For the purpose of integrating the components of the argument structure, we split the 31 available topics into 24 topics for training, 4 for tuning hyper-parameters and determining early stopping (dev set), and 3 for testing, in line with previous work \citep{DBLP:conf/acl/Bar-HaimEFKLS20}. The data distribution is shown in Appendix~\ref{sec:appendix_data_dist}. In total, 2002 texts in Task 1 were considered relevant to some claim, of which 1456 were labelled as evidence; the number is higher than the 1881 unique evidence pieces, because each piece can support support multiple claims. 
For Task 2, two pieces of evidence for the same claim were paired together and a total of 6034 pairs were obtained. The annotation results were quantified into three labels based on the degree of convincingness of the former over the latter, i.e. ``1'' if the former is more convincing than the latter, ``0'' vice versa, and ``0.5'' when both are equally convincing. As shown in Table \ref{tab:example_alloftasks}, in Task 3, claims for the same stance on a topic and the evidence supporting them were then summarised into a debate script. Generated by each of the six previously mentioned models according to the six settings, a total of 2232 unique summaries were obtained. The summaries were then ranked by humans along the evaluation dimensions mentioned in Task 4, resulting in 1488 pairwise rankings.


\textbf{Randomness in Crowdsourcing Annotation Platforms and Spam Filtering Mechanism:} AMT is a widely recognized and extensively used crowdsourcing annotation platform \citep{paolacci2010running}, but ensuring the quality of annotators is challenging \citep{aruguete2019serious}. Consequently, we implemented two post-processing measures. On the one hand, to monitor and ensure the quality of the annotations collected, the following measures were employed: (1) \textbf{Test questions:} Inspired by \citet{DBLP:conf/emnlp/ToledoGCFVLJAS19}, a unique set of 100 hidden test questions was designed as an attention check, where one of the options were other text snippets from unrelated topics. These test questions were presented in the same way as the other questions. The work of annotators who failed 20\% of the test questions was ignored. (2) \textbf{Annotator reliability testing:} 4\% (250) of the annotations were selected for judgement by experts. For each annotator, the \emph{average agreement} score with other annotators was calculated on the data they annotated, and then the average agreement of all annotators were ranked. If an annotator's average agreement differed too much (lower than 0.125) from the average agreement scores of other annotators, they were not considered for follow-up annotation\footnote{We release raw annotation data, to facilite subsequent research to estimate gold standards from the noisy crowdsourced labels \citep{DBLP:conf/emnlp/Plank22}}.

\textbf{ChatGPT as Third Annotator in Task 2:} Previous work has demonstrated that ChatGPT can already achieve performance comparable to that of crowdsourcing annotators \citep{DBLP:journals/corr/abs-2303-15056}. Based on this, we first verified the reliability of ChatGPT annotations by annotating all of the test instances mentioned in \textbf{Annotator reliability testing}. ChatGPT's results were compared with golden labels as well as instances where human annotators reach agreement separately. This ranked ChatGPT third out of 124 human annotators who participated in annotating for Task 2 and reaching a 97\% agreement with human annotators in the latter. Thus, ChatGPT was employed as a ``third annotator'' for all instances that were annotated by an even number of human evaluators with no preference for the results. 
If ChatGPT's answer yielded no preference (i.e. \emph{``Both evidence 1 and evidence 2 support the claim...''}), we considered both pieces of evidence to be equally persuasive and assigned the label ``0.5''.

\begin{table}[t]
    \centering
    \scalebox{0.52}{
    \begin{tabular}{lcccc:cccc}
         \toprule
         Approach & HE & TR & HM & FL & SG & IF & FT & RE \\
         \midrule
         \emph{${\rm GPT4}$} & 4.6388 & 4.8888 & 4.9722 & 4.6944 & 4.8611  & 4.5278 & 4.9444 & 4.9722  \\
         \emph{${\rm GPT3.5}$} & 4.5104 & 4.5555 & 4.8750 & 4.5729 & 4.4861  & 4.4409 & 4.6319 & 4.5798 \\
         \emph{${\rm GPT3}$} & 4.3715 & 4.4027  & 4.8472 & 4.4652 & 4.3333 & 4.3125 & 4.5138 & 4.4201  \\
         \emph{${\rm Bard}$} & 4.3402 & 4.4513 & 4.8333 & 4.4618 & 4.2083  & 4.2048 & 4.5312 & 4.3923  \\
         \emph{${\rm Alpaca_{13B}}$} & 4.3958 & 4.5381 & 4.8750 & 4.4826 & 4.5555  & 4.3819 & 4.5937 & 4.5277  \\
         \emph{${\rm Vicuna_{13B}}$} & 4.3899 & 4.4174 & 4.8348 & 4.4633 & 4.3211  & 4.2981 & 4.5045 & 4.4403  \\
         \bottomrule
    \end{tabular}
    }
    \caption{Performance of different approaches on each dimension in human evaluation. Each score is averaged over five annotators on the dimension. Reported are, from left to right, \textsc{Helpful}, \textsc{Truthful}, \textsc{Harmless}, \textsc{Fluent}, \textsc{Significant}, \textsc{Informative}, \textsc{Faithful} and \textsc{Redundant}.}
    \label{human_result}
\vspace{-4mm}
\end{table}

\textbf{Reliability of Human Annotation:} Overall, a total of 1636 human annotators participated in our annotation project. Considering that the evidence was collected after filtering by human experts, the difficulty of Task 1 was low. We hired six experts for this annotation task who participated in an onboarding training with 250 data samples. Finally, four annotators participated in the whole annotation process. We measured Cohen's Kappa \citep{mchugh2012interrater} to assess inter-annotator agreement, reporting an average of 0.6886 across the train/dev/test set, implying that the results are substantially reliable 
\citep{munoz1997interpretation}. Agreement on Task 2 was significantly lower, presumably because the task depended on the participants' background and knowledge. Furthermore, a different set of annotators for each annotation 
resulted in difficulty to obtain meaningful values on Cohen's Kappa. 
Therefore, we measure reliability by comparing the pairwise agreement between each two annotators who annotated at least one example together, reporting such an an average agreement of 0.549. We further used ChatGPT with verified reliability as an additional annotator to resolve tie breaks, as mentioned in the previous section. For Task 3, we did a cross-validation for every two annotators that had 25\% overlap between their work, resulting in Krippendorff's $\alpha$ \citep{krippendorff2011computing} of 0.2474, which suggests that providing rankings is a largely subjective task. 
For Task 4, Cohen's Kappa was measured at 0.8970 for sufficiency and 0.4933 for factuality, respectively. These suggest that the task of identifying factual errors and evaluating argumentative logic within the generated text poses a considerable challenge. Conversely, determining the persuasiveness of debate scripts appears straightforward (see Appendix~\ref{sec:appendix_human_annotators} for details of the annotation platform, costs and templates).

 \begin{table*}
\centering
\scalebox{0.9}{
    \begin{tabular}{lcccc}
   \toprule
   \multirow{2}{*}{Systems} & \multicolumn{3}{c}{R-1/R-2/R-L} & BLEURT\\
    \cmidrule(r){2-4}
   {} & Best-Evi & Top2-Evi & All-Evi &  Best/Top2/All\\
   \midrule
   \textbf{Golden}  & {} \\
   \emph{${\rm GPT3.5}$} & 46.58/18.12/22.98 & 55.41/25.65/25.10 & 50.00/14.57/25.81 & 45.91/44.61/39.27 \\
   \emph{${\rm GPT3}$} & 41.21/12.8/19.39 & 51.58/20.04/25.3 & 48.22/13.09/20.15 & 42.87/48.19/46.73 \\
    \emph{${\rm Bard}$} & 32.21/9.3/17.26 & 44.14/15.57/21.30 & 42.74/11.2/22.65 & 43.07/43.52/51.70 \\
   \emph{${\rm Alpaca_{13B}}$} & 14.28/2.4/10.31 & 18.51/2.97/11.48 & 30.73/4.27/16.54 & 36.19/42.95/49.40 \\
   \emph{${\rm Vicuna_{13B}}$} & 51.08/19.56/28.13 & 41.86/20.65/27.33 & 41.61/13.58/17.37 & 34.87/37.49/31.54\\
   \midrule
   \textbf{Baseline}  & {} \\
   \emph{${\rm GPT3.5}$} & 57.31/28.83/37.19 & 55.13/26.08/32.43 & 55.12/29.41/34.14 & 51.67/48.66/52.01\\
   \emph{${\rm GPT3}$} & 53.46/21.26/29.04 & 46.98/19.8/26.66 & 54.99/17.15/26.25  & 47.14/48.50/44.38 \\
    \emph{${\rm Bard}$} & 41.53/17.16/23.38  & 40.99/17.13/21.42 & 42.45/12.28/19.65 & 50.08/43.05/45.67\\
   \emph{${\rm Alpaca_{13B}}$} & 41.25/10.69/21.25 & 30.00/4.3/15.71 & 26.38/7.9/18.05 & 48.97/44.33/48.83\\
   \emph{${\rm Vicuna_{13B}}$} & 56.86/31.88/38.82 & 50.68/25.77/29.10 & 44.02/19.06/22.56 & 46.11/43.39/44.80\\
   \bottomrule
\end{tabular}
}
    \caption{Performance of the proposed method compared with GPT4 generation as reference, measured by \{1,2\}-gram overlap (Rouge-\{1,2\}) and longest sequence overlap (Rouge-L). Best, Top2 and All represent different evidence settings.}
    \label{TASK3B}
\end{table*}

\section{Baseline Approaches}
\textbf{Sentence-pair Classification:}
We formulate Tasks 1, 2 and 4 as sentence-pair classification tasks, where we employed \texttt{Flan-T5} \citep{DBLP:journals/corr/abs-2210-11416}, \texttt{BERT} \citep{devlin2018bert} and \texttt{RoBERTa} \citep{liu2019roberta} as baseline models. As an abstract representation of the task \emph{''Does the evidence support the claim''}, the input is a concatenation of argument and evidence candidate, and the output is binary, i.e., $\{1, 0\}$, where the snippet is considered as evidence for the claim when they are judged to be the same pair. Task 4 (sufficiency and factual error checking, where $\{1, 0\}$ represent sufficient/factually wrong or not, respectively) follows the same setup. For Task 2, we define it as a multi-class classification task, i.e., given an evidence pair and a claim, ``the former is more convincing'' is labelled as 1, ``the latter is more convincing'' is labelled as 0, and ``both are equal'' is labelled as 0.5.

\textbf{LLMs with Contrastive Learning:}
We formulate Tasks 2 and 3 as a contrastive learning task. Conceptually, we consider the evidence closest to the claim in the semantic embedding space as the most supportive candidate for it. For this purpose, we utilise Siamese Neural Networks (including \texttt{SimCSE-RoBerta} \cite{gao2021simcse}, \texttt{Sentence-t5} \citep{ni2021sentence}) with contrasting loss functions. The inputs are the concatenation of the topic and claim, and the evidence candidate, respectively. The output is the score reflecting how close they are to each other (more details in Appendix~\ref{sec:appendix_cl}).

\textbf{Summarisation with LLMs:}
Inspired by previous work in argument summarisation \citep{DBLP:conf/acl/LiSBN23}, we employed Flan-T5 \citep{DBLP:journals/corr/abs-2210-11416} and PEGASUS \citep{DBLP:conf/icml/ZhangZSL20} as summarisation systems. The former is a large model trained on more than a thousand instruction tasks, while the latter has pre-trained objectives tailored specifically for abstract summarisation. The input of the model is in the form of a concatenated pair \emph{<stance, topic, claim 1, evidence 1, evidence 2,..., claim 2, ....>}. The output will be a single summary.

\textbf{Project Debater:}
The first end-to-end automatic debating system, Project Debater \citep{slonim2021autonomous} encompasses multiple components such as claim detection, evidence detection, argument quality assessment and narrative generation. Employing it on our dataset, we can explore the limitations of a component-based SotA system.

\section{Experimental Settings}

Broadly speaking, we aim to investigate how well the generated data aligns with human standard and the performance of current representative LLMs in end-to-end debate preparation; we also provide a benchmark for future research. Specifically, we ask: \emph{(i)} How well do representative LLMs perform on tasks? \emph{(ii)} How well does LLM-generated data correlate with human judgement? \emph{(iii)} How big is the impact of end-to-end pipeline inputs on model performance? \emph{(iv)} How good is the baseline performance on each of the subtasks? To answer questions  (\emph{i}) and (\emph{ii}), we conduct both manual evaluation and automated evaluation by first evaluating LLM outputs by means of human evaluation and then assessing correlations with human judgement and learned metrics. For questions \emph{(iii)} and \emph{(iv)}, we compare the performance of our proposed baseline approach with both golden input and pipeline input.

 To measure the performance on the imbalanced data distribution setting, we measure the Matthews correlation coefficient (MCC) \citep{guilford1954psychometric} as well as Macro-F1. The former progressively penalises over-confidence on the majority class, while the latter allocates the same weight to each class regardless of its cardinality in the dataset. For the summarisation task, we employ the widely used ROUGE metric~\cite{lin2004rouge}, which measures performance by counting $n$-gram overlap. In addition, BLEURT~\citep{DBLP:conf/acl/SellamDP20} is used to measure the performance from a semantic similarity perspective. For the ranking task, we consider precision@$k$,  Mean Average Precision@$k$ (MAP@$k$) \cite{davis2006relationship} and Normalized Discounted Cumulative Gain@$k$ (NDCG@$k$) \cite{burges2005learning}. MAP@$k$ is  concerned with precision at every rank up to $k$, which takes into account both the number of relevant documents retrieved and their order in the rank list, penalising systems that return relevant documents further down the list. Meanwhile, NDCG gives more weight to the correctness of items at higher positions in the ranked list. 
 Both metrics are widely used to evaluate ranking systems. Further details on parameter settings of baselines can be found in Appendix~\ref{sec:experiment_setting}.

\section{Benchmarks Results and Analysis}




\textbf{LLMs Generation Aligns with Human Standard, but Weaknesses in Logic and Informativeness persist:} Table~\ref{human_result} shows that human evaluations suggest that generated summaries are of high quality. Overall, GPT4 showed the best performance, while Alpaca had the worst performance which is consistent with the results of the BT model based on ranked quantification (shown in Appendix \ref{sec:bt_irt}). Specifically, all models perform well in the \textsc{Truthful}, \textsc{Harmless}, \textsc{Faithful} and \textsc{Redundant} categories, suggesting that LLM outputs are comparable to humans along those dimensions. However, they still lack in dimensions that require logical understanding (\textsc{Helpful} and \textsc{Informative}).
                                         
\textbf{Similarity to GPT-generated Summaries Correlates with Argumentative Quality Assessments:} The relative quality of the models is further represented in Table \ref{TASK3B} and \ref{TASK3Bt_ci}. Using human-evaluated performance of GPT-4 as a reference, ChatGPT performs the best, while Alpaca's performance drops the most. We further explore the correlation between the human evaluation with Rouge scores (Spearman's $r=0.75, p<.005$) and BLEURT (Spearman's $r=0.64, p<.005$), suggesting that measuring the similarity of outputs to GPT4-generated summaries is a feasible proxy-metric representing the evaluation of argument quality. Intuitively, this is possible as GPT4 was assigned nearly full marks in all dimensions during human evaluation. Therefore, we treat its output as the gold standard for grammatical dimensions (e.g., \textsc{Fluent} and \textsc{Informativeness} \citep{zhao2023survey}) in the following experiment that investigates the performance of baseline approaches. 
Nevertheless, quality assessment still relies on human annotators.

\begin{table}[htbp]
\centering
\scalebox{0.63}{
\begin{tabular}{lcc}
   \toprule
    \multirow{2}{*}{Systems} & Rouge & BLEURT\\
   {} & Golden/Baseline & Golden/Baseline\\
   \midrule
   \emph{${\rm GPT3.5}$} & $24.63 \pm 2.99$/$34.59 \pm 4.89$ & $43.26 \pm 7.14$/$50.78 \pm 3.74$ \\
   \emph{${\rm GPT3}$} & $21.61 \pm 6.51$/$21.61 \pm 6.51$ & $45.93 \pm 5.58$/$46.67 \pm 4.26$ \\
    \emph{${\rm Bard}$} & $20.40 \pm 5.69$/$21.48 \pm 3.78$ & $46.09 \pm 9.85$/$46.26 \pm 7.21$ \\
   \emph{${\rm Alpaca_{13B}}$} & $12.77 \pm 6.72$/$18.34 \pm 5.64$ & $42.85 \pm 13.40$/$47.37 \pm 5.35$ \\
   \emph{${\rm Vicuna_{13B}}$} & $24.27 \pm 12.16$/$30.16 \pm 16.60$ & $34.63 \pm 6.05$/$ 44.77 \pm 2.76 $ \\
   \bottomrule
\end{tabular}
}
    \caption{Performance of the proposed method compared with human-evaluated generation of GPT-4 as a reference, measured by Rouge-L sequence overlap and BLEURT at CI 95\%}
    \label{TASK3Bt_ci}
\end{table}


\begin{table}
\centering
\scalebox{0.9}{
\begin{tabular}{lcc}
   \toprule
   Systems & Rouge & BLEURT\\
   \midrule
   \textbf{Golden}  & {} \\
   \emph{${\rm FlanT5_{3B}}$} & $39.62 \pm 9.94$ & $43.26 \pm 7.14$ \\
   \emph{${\rm FlanT5_{Large}}$} & $37.18 \pm 7.92$ & $45.93 \pm 5.58$ \\
    \emph{${\rm PEGASUS}$} & $24.12 \pm 9.87$ & $46.09 \pm 9.85$ \\
   \midrule
   \textbf{Pipeline}  & {} \\
    \emph{${\rm FlanT5_{3B}}$} & $24.54 \pm 2.73$ & $34.63 \pm 6.05$\\
   \bottomrule
\end{tabular}
}
    \caption{Performance of the proposed method on Task 3 subtask A, measured by Rouge-longest sequence overlap with GPT4's output (both golden and baseline settings) as reference and confidence interval at 95\% }
    \label{task3a_baselinet_ci}
\end{table}

\textbf{End-to-end Performance Exhibits Significant Drops, Demonstrating the Challenges of the Task:} We select the most powerful model in each of subtask to build the end-to-end pipeline, where the input of each subsequent task is the output of the preceding task. As can be seen from Table~\ref{task2_baseline}, the performance of the pipeline has a significant drop compared to the same model with golden inputs, dropping $0.11$ points Prec@$1$ on Task 2 and $16$ points ROUGE on Task 3, respectively. This shows that achieving an integrated debating system that is fully aligned with human preferences is challenging. 
It should be noted that Project Debater also shows a significant performance drop, indicating that there is still room for further improvement of end-to-end systems, underlining the importance of our collected resources. 


\textbf{Task 1: Evidence Detection} Overall, \texttt{FlanT5-11B} achieved the best performance, 
as shown in Table \ref{task1_baseline}. Specifically, \texttt{FlanT5-11B} and \texttt{FlanT5-3B} perform significantly better than smaller models. 
This suggests that larger models perform better when faced with label imbalance. \texttt{FlanT5} also outperforms the encoder-only models of the same size, suggesting that either instruction-finetuning or the encoder-decoder architecture improve performance. 
Regarding specific classes, all models performed better in predicting evidence, likely because it is the majority label. In contrast to performance reported so far, \texttt{RoBerta-Large} performed best (with Accuracy of $0.52$) when predicting non-evidence, and correspondingly it performed worst with an Accuracy of $0.79$ when predicting evidence, indicating that it achieved a relatively balanced overall performance. Finally, compared to inference (zero-shot), the models trained on the proposed dataset always perform better, suggesting that the dataset can effectively improve the performance of the models in this task. 
\texttt{FlanT5-3B} shows comparable overall performance with Project Debater; bigger LLMs could potentially lead to improved performance.

\begin{table}[htbp]
\centering
\scalebox{0.6}{
    \begin{tabular}{lccccc}
   \toprule
   Systems & Acc & MCC & Macro-F1 & Acc-Evi & Acc-NonEvi \\
   \midrule
   \emph{${\rm FlanT5_{11B}}$} & 0.8430 & 0.4030 & 0.6827 & 0.9592 & 0.3478 \\
   \emph{${\rm FlanT5_{3B}}$} & 0.8306 & 0.3906 & 0.6899 & 0.9286 & 0.4130 \\  
   \emph{${\rm FlanT5_{Large}}$} & 0.8099 & 0.3409 & 0.6687 & 0.9031 & 0.4130 \\ 
   \emph{${\rm RoBerta_{Large}}$} & 0.7520 & 0.2709 & 0.6282 & 0.7857 & 0.5217 \\ 
   \emph{${\rm Bert_{Large}}$} & 0.7606 & 0.1839 & 0.5873 & 0.8979 & 0.2608 \\ 
   \hdashline
   \textbf{Zero-shot}  & {}  \\
   \emph{${\rm Project Debater}$} & 0.5219 & 0.0433 & 0.5215 & 0.5407 & 0.5025\\
   \emph{${\rm FlanT5_{3B}}$} & 0.5102 & 0.1489 & 0.6123 & 0.4653 & 0.7317 \\
   \bottomrule
\end{tabular}
}
\caption{Matthews correlation coefficient of the proposed method on Task 1 Evidence Detection.}
\label{task1_baseline}
\end{table}

\begin{table}
\centering
\scalebox{0.6}{
    \begin{tabular}{lcccc}
   \toprule
   Methods & Prec@1 & Prec@2 & MAP@2 & NDCG@2 \\
   \midrule
   \emph{${\rm Project Debater}$} & 0.5022 & 0.606 & 0.5044 & 0.5740 \\
   \emph{${\rm Classification (CF)}$} & 0.3939 & 0.5 & 0.4394 & 0.5220 \\
   \emph{${\rm Contrastive\ Learning (CL)}$} & 0.2727 & 0.3939 & 0.3560 & 0.4194 \\
   \hdashline
   \textbf{Pipeline}  & {}  \\
   \emph{${\rm Project Debater}$} & 0.4244 & 0.5156 & 0.4544 & 0.5156  \\
   \emph{${\rm FlanT5_{CLS}}$} & 0.1818 & 0.3939 & 0.3257 & 0.3939 \\
   \bottomrule
\end{tabular}
}
\caption{Performance of the proposed method on Task 2. Pipeline represents the input from last components. Key: \texttt{Prec@k} =  precision at top-k, \texttt{MAP} = Mean Average Precision, \texttt{NDCG} = Normalized Discounted Cumulative Gain}
\label{task2_baseline}
\end{table}

\textbf{Task 2: Evidence Convincingness Ranking} As mentioned in the baseline setting description, Task 2 has been formulated as both a classification and ranking task. Comparing the performance of classification vs contrastive learning, the former outperforms the latter~(Table~\ref{task2_baseline}). 
This indicates  that the label ``0.5'' (the same persuasive power for both) bears important semantics, which is lost incontrastive learning, resulting in lower scores. In detail, it is noticeable that model size does not seem to play a significant role for the contrastive learning approach (c.f. SentenceT5-Large and -3B in Table~\ref{task2b_baseline}) 
which might be because the nuanced differences in evidences are hard to pick up by means of semantic similarity. For classification methods, performance scales with size, which is consistent with Task 1 results. 

\begin{table}[hp]
\centering
\scalebox{0.6}{
    \begin{tabular}{lcccc}
   \toprule
   Systems & Acc & Macro-F1 & Acc-Con & Acc-NonCon \\
   \midrule
   \textbf{Classification}  & {}  \\
   \emph{${\rm FlanT5_{3B}}$} & 0.5232 & 0.3367 & 0.2907 & 0.8071 \\
   \emph{${\rm FlanT5_{Large}}$} & 0.4118 & 0.2890 & 0.2056 & 0.5814 \\
   \hdashline
   \textbf{Contrastive Learning}  & {}  \\
   \emph{${\rm SentenceT5_{3B}}$} & 0.5528 & 0.5565 & 0.5254 & 0.5843 \\
   \emph{${\rm SentenceT5_{Large}}$} & 0.5461 & 0.5621 & 0.5457 & 0.5465 \\ 
   \emph{${\rm SimCSE}$-${\rm RoBERTa_{Large}}$} & 0.5569 & 0.5465 & 0.5000 & 0.6221 \\
   \bottomrule
\end{tabular}
}
\caption{Detailed performance of the proposed method on Task 2 Evidence Convincingness Ranking }
\label{task2b_baseline}
\end{table}

 \begin{table*}
\centering
\scalebox{0.9}{
    \begin{tabular}{lcccc}
   \toprule
   \multirow{2}{*}{Systems} & \multicolumn{3}{c}{R-1/R-2/R-L} & BLEURT  \\
    \cmidrule(r){2-4}
   {} & Best-Evi & Top2-Evi & All-Evi &  Best/Top2/All  \\
   \midrule
   \textbf{Golden}  & {} \\
   \emph{${\rm FlanT5_{3B}}$} & 59.20/32.34/45.08 & 53.10/26.16/35.59 & 55.55/28.03/38.21 & 53.21/53.52/49.96\\
   \emph{${\rm FlanT5_{Large}}$} & 55.93/29.97/41.69 & 51.66/25.55/35.12 & 51.97/25.90/34.75 & 31.20/4927/45.24\\
   \emph{${\rm PEGASUS}$} & 39.06/21.38/27.25 & 38.75/19.62/26.60 & 26.00/9.74/18.52 & 28.14/29.50/34.84\\
   \midrule
   \textbf{Pipeline}  & {} \\
   \emph{${\rm FlanT5_{3B}}$} & 42.18/13.81/23.56 & 44.95/17.18/23.99 & 47.77/19.29/26.08 & 44.32/45.96/43.10\\
   \bottomrule
\end{tabular}
}
    \caption{Performance of the proposed method on Task 3 subtask A, measured by \{1,2\}-gram overlap Rouge-\{1,2\}, longest sequence overlap Rouge-L with GPT4's output (both golden and baseline settings) as reference. Best, Top2, All represent different evidence settings.}
    \label{task3a_baseline}
\end{table*}

\textbf{Task 3 Subtask A: Argumentation Summary} Here, \texttt{FlanT5-3B} outperformed all other models in all settings, with the remaining models exhibiting comparable performance levels (c.f., Average ROUGE scores of \emph{All} vs \emph{Top-2} in Table~\ref{task3a_baseline}). This suggests that this particular setting presents the most challenges, potentially due to constraints in input length and the model's struggle to select relevant evidence.  It is worth noting that there is a huge gap between the performance of the model with end-to-end inputs and the model with golden inputs, which demonstrates the challenges of the task (Shown in Table \ref{task3a_baselinet_ci}). 


\textbf{Task 3 Subtask B: Argumentation Summary and Ranking} Table~\ref{task3b_baseline} shows the performance on the proposed baseline models, suggesting that the size of the model does not significantly contribute to succeeding on this task. 
However, \texttt{SentenceT5} performs better than \texttt{SimCSE-RoBERTa-Large}, 
 suggesting that embedding two sentences simultaneously has better performance in measuring similarity than embedding sentences separately, presumably due to the fact that the former is better at measuring the relationship between the original and interfering inputs by representing them in the same space than what the latter does, i.e., by representing them separately \citep{neculoiu2016learning}. 

\begin{table}[htbp!]
    \centering
    \scalebox{0.65}{
        \begin{tabular}{lcccc}
        \toprule
        Systems & Prec@1 & Prec@2 & MAP@2 & NDCG@2 \\
        \midrule
        \emph{$\rm SentenceT5_{3B}$} & 0.25 & 0.5 & 0.375 & 0.4811 \\
        \emph{$\rm SentenceT5_{Large}$} & 0.25 & 0.5278 & 0.4097 & 0.5089 \\
        \emph{${\rm SimCSE}$-${\rm RoBERTa_{Large}}$} & 0.25 & 0.4722 & 0.3819 & 0.4722 \\
        \bottomrule
        \end{tabular}
    }
    \caption{Performance of the proposed method on Task 3 Subtask B: Argumentation Summary and Ranking (ASR), \texttt{Prec@k} represents precision at top-k, \texttt{MAP} represent Mean Average Precision, \texttt{NDCG} represent Normalized Discounted Cumulative Gain.}
    \label{task3b_baseline}
\end{table}

\textbf{Task 4: Summary Quality Checking}
Considering that data sets are imbalanced for both tasks, the smaller models default to predicting the majority class resulting in an MCC score of $0$. \texttt{FlanT5-3B} is the only model that performs marginally better 
albeit with only weak correlation to human judgements. This suggests that the large model exhibits enough robustness to address the challenges posed by imbalanced data and the weak human feedback signal. Specifically, the task of fact-checking appears to be particularly challenging, as all the large-sized models exhibit an MCC of $0$. This could be attributed to the scarcity of data labeled as ``1'', i.e. factually wrong.

\begin{table}[h]
\centering
\scalebox{0.65}{
    \begin{tabular}{lccccc}
   \toprule
   Systems & Acc  & MCC & Macro-F1 & Acc-True & Acc-False \\
   \midrule
   \textbf{Sufficiency}  & {}  \\
   \emph{${\rm FlanT5_{3B}}$} & 0.8819 & 0.2890 & 0.6136 & 0.9802 & 0.1944\\
   \emph{${\rm FlanT5_{Large}}$} & 0.8472 & 0.1446 & 0.5648 & 0.9444 & 0.1667\\
   \emph{${\rm RoBerta_{Large}}$} &  0.8101 & 0.0000 & 0.4475 & 1.0 & 0.0\\
   \emph{${\rm Bert_{Large}}$} & 0.8101 & 0.0000 & 0.4475 & 1.0 & 0.0\\
   \hdashline
   \textbf{Faithfulness}  & {}  \\
   \emph{${\rm FlanT5_{3B}}$} & 0.9236 & 0.3328 & 0.5871 & 0.1200 & 1.0000\\
   \emph{${\rm FlanT5_{Large}}$} & 0.9027 & 0.0000 & 0.4744 & 0.0000 & 1.0000\\
   \emph{${\rm RoBerta_{Large}}$} & 0.9027 & 0.0000 & 0.4744 & 0.0000 & 1.0000\\
   \emph{${\rm Bert_{Large}}$} & 0.9027 & 0.0000 & 0.4744 & 0.0000 & 1.0000\\
   \bottomrule
\end{tabular}
}
\caption{Performance of the proposed method on Task 4. Sufficiency represents the sufficiency checking, while faithfulness represents the fact-checking task.}
\label{task4_baseline}
\end{table}


\section{Conclusion}
In this study, we built an end-to-end argumentative summary and evaluation dataset and established the benchmarks for each task. This process involves annotating the data by human experts mixed with LLMs and applying popular deep learning baseline methods to the new datasets. We open-source our data and benchmark code, enabling researchers to reproduce and enhance the results. In future, we hope to further expand the evaluation dataset to explore the potential of unprocessed real-world data as evidence and fact-checking for its quality.

\section*{Limitations}
Recruiting human subjects for annotation limits the reproducibility of human evaluation. In addition, we recognise that there might be more suitable baseline models, and in this study, just a few of the most advanced models were used.  Also, we may employ human experts to complete summaries that are not generated as expected for different reasons to fill in the gaps. Finally, despite being filtered by human experts, offensive debate content may still be present in the data. It is worth noting, however, that the identification of offensive language is not the aim of this work.

\section*{Ethics Statement}
During the annotation process, we considered the following dimensions to protect our annotators. (1) Consent: To ensure that our participants, i.e., AMT workers, consent to the annotation task, they will be required to review the task guidelines and instructions before starting their work. They can exit the task at any time if they feel uncomfortable. (2) Confidentiality: None of the task participants is known to us, i.e., the entire annotation process is anonymised: no information about the AMT workers is accessible to us, and none of the data being annotated is confidential as the data samples all come from publicly available sources. (3) Safeguarding: We selected debate topics (for annotation) that are used in high school-level debates and argumentative essays. They are unlikely to cause any harmful effects, as they have already been subject to previous scrutiny (Released by KPA dataset \citep{DBLP:conf/acl/Bar-HaimEFKLS20}).

\section*{Acknowledgements}
We thank the anonymous reviewers from the ARR December 2023 cycle for their valuable feedback. We would especially like to thank Roy Bar-Haim and Yoav Katz at the IBM Research for helpful discussions, feedback on our work and help with using Project Debater. We would also like to acknowledge the use of the Computational Shared Facility at The University of Manchester. We would like to conclude by thanking all of our hard-working annotators. This work was partially funded by the European Union’s Horizon 2020 research and innovation action programme, via the AI4Media Open Call \#1 issued and executed under the AI4Media project (Grant Agreement no. 951911).

\bibliography{anthology,custom}

\appendix

\section{Milp Ranker}
\label{sec:appendix_milp_ranker}
The \texttt{milp\_ranker} formulates the problem that ranking from pairwise comparisons as solving a mixed-integer linear program (MILP). We select it as our ranker because it can handle both the contradictory case between the pairwise comparisons and the equal case between two pieces of evidence. The use of \texttt{milp\_ranker} is restricted and therefore an academic license was applied here. To align with annotated results in Task 1, evidence with the label 0 in Task 1 was removed from the ranking.

\section{Quality evaluation dimensions for summaries}
\label{sec:quality_summaries}

\begin{itemize}
    \item HELPFUL: The output should follow the user’s intention and help users prepare a script for debate that meets their demands
    
    \item TRUTHFUL: The output contains accurate information and doesn’t mislead the user.
    
    \item HARMLESS: The output should not cause physical, psychological, or social harm to people; or harm to institutions or resources necessary to human wellbeing.
    
    \item SIGNIFICANT: Each claim and supporting evidence(s) should stand out and capture a main point.
    
    \item INFORMATIVENESS: It should discuss some aspect of the debate topic and be general enough. Any claims or evidence(s) that are too specific or only express sentiment cannot be considered a good candidate.
    
    \item FAITHFULNESS: Claims and Evidence should actually express the meaning in the corpus. No conjecture or unfounded claims arise.
    
    \item REDUNDANT: Each claim expresses a distinct aspect. In other words, there should be no overlap between the claims.

    \item FLUENT: tthe generated text constitutes a fluent script that can be presented by humans.
\end{itemize}

\section{Bradley–Terry model (BT) and Item Response Theory (IRT)}
\label{sec:bt_irt}
Here, we formalise the BT model into the following:

\begin{align}
    {\displaystyle L(\mathbf {p} )=\ln \prod _{ij}[P(i>j)]^{w_{ij}}=\sum _{ij}w_{ij}\ln {\frac {p_{i}}{p_{i}+p_{j}}}}
\end{align}

\begin{align}
    {\displaystyle p_{i}={\frac {\sum _{j}w_{ij}p_{j}/(p_{i}+p_{j})}{\sum _{j}w_{ji}/(p_{i}+p_{j})}}}
\end{align}

where $p_{i}$ is a positive real-valued score assigned to individual $i$, $w_{ij}$ be the number of times individual $i$ beats individual $j$. 

Based on the fact that each annotator's preference may differ from others as well as the diversity of the models' generation, we normalise the performance of the model into scores using the two-parameter logistic model following the Item Response Theory (IRT) \cite{embretson2013item}. Here, we formalise the task into the following:

 \begin{equation}
     P(X_{ij} \arrowvert \theta_{i}; \alpha_{j}; \beta_{j}) = \frac{e^{D\alpha_{j}(\theta_{i}-\beta_{j})}}{1+e^{D\alpha_{j}(\theta_{i}-\beta_{j})}}
 \end{equation}


 where $\theta_{i}$ represent the ability of the $i$th models , $\alpha_{j}$ represent the discrimination of the $j$ topic, $\beta_{j}$ represent the difficulty of the $j$ topic, $X_{ij}$ represent the $i$th model's response matrix on the topic $j$ and $D$ is a constant and takes the default value of 1.702, where the probability density of the function differs from the normal shoulder curve by less than 0.01.
 
  \begin{table}[t]
\centering
\scalebox{0.75}{
    \begin{tabular}{lcccc}
   \toprule
   Systems & Best-Evi & Top2-Evi & All-Evi & Average \\
   \midrule
   \textbf{Golden Setting}  & {} \\
   \emph{${\rm GPT4}$} & 2.8905 & 2.8906 & 2.5164 & 2.7658 \\
   \emph{${\rm GPT3.5}$} & 1.5718 & 0.4122 & 0.9578 & 0.9806 \\
   \emph{${\rm GPT3}$} & -0.7382 & -2.0713 & 0.9578 & -0.6172 \\
    \emph{${\rm BARD_{340B}}$} & 0.4123 &  1.5718 & -0.3021 & 0.5607 \\
   \emph{${\rm Alpaca_{13B}}$} & -2.0712 & -2.0713 & -1.4443 & -1.8622 \\
   \emph{${\rm Vicuna_{13B}}$} & -2.0712 & -0.7382 & -2.6917 & -1.8337 \\
   \hdashline
   \textbf{Baseline Setting}  & {} \\
   \emph{${\rm GPT4}$} & 3.1028 & 2.3864 & 2.8905 & 2.7932 \\
   \emph{${\rm GPT3.5}$} & -0.5643 & 1.2032 & 0.4122 & 0.3504 \\
   \emph{${\rm GPT3}$} & 1.7379 & -0.1 & 1.5717 & 1.0699 \\
    \emph{${\rm BARD_{340B}}$} & 0.5623 & -0.1 & 0.7381 & 0.4001 \\
   \emph{${\rm Alpaca_{13B}}$} & -3.1048 & -2.3884 & -2.0711 & -2.5214 \\
   \emph{${\rm Vicuna_{13B}}$} & -1.7399 & -1.2052 & -2.0711 &  -1.6721\\
   \bottomrule
\end{tabular}
}
\caption{Performance of different systems on Task 3 summary generation reflect by IRT model, different evidence inputs selected by both human annotators’ results and proposed baseline model separately. A higher probability of the keying response represents the better performance of the model on that task and vice versa.}
\label{task3_irtresult}
\end{table}

\section{Dataset Distribution}

Table \ref{dataset_analysis} Demonstrate the distributions of the dataset. For Task 1 Evidence Detection, numbers in brackets represent the number of cases labelled as evidence.
\label{sec:appendix_data_dist}
\begin{table}[htbp]
\centering
\scalebox{0.7}{
    \begin{tabular}{lccccc}
   \toprule
   Task & ED & ECR & AS & ASR & SQE \\
   \midrule
   Train & 1494(1033) & 4300 & 1728  & 1152 & 1728\\
   Dev & 266(227)  & 935 &  288 & 192 & 144\\
   Test & 242(196) &  799  &216 & 288 & 216\\
   \bottomrule
\end{tabular}
}
\caption{The distribution of the dataset, ED represents Evidence Detection, ECR represents Evidence Convincingness Ranking, AS represents Argumentation Summary, ASR represents Summary Quality Ranking and SQE represents Summary Quality Checking.}
\label{dataset_analysis}
\end{table}

\section{Human Annotators Recruited and Cost}
\label{sec:appendix_human_annotators}
\subsection{Overview of Annotation Process}
We release the annotation task and employ annotators from the Amazon Mechanical Turk crowdsourcing platform directly. The project spends \pounds3160 in total, where each of the data cost \$0.2. Any annotator can start working after carefully reading the instructions and can choose to quit at any time.

\subsection{Annotation Template}

We provide a annotation template that we shape the task into bianry classification task in order to simplify the task, example can be seen in Table \ref{annotation_task1}. And annotation template for summarisation ranking task in Table \ref{annotation_task3}.
\label{sec:template}
\begin{table*}[htbp]
    \centering
    \scalebox{1}{
        \begin{tabular}{p{10cm}p{5cm}}
        \toprule
        Example Annotate Template For Task 1 & {} \\
        \midrule
        \textbf{Input Instance} & {} \\
        Evidence Candidate: Research by the Schoolwear Association found that 83\% of teachers thought a good school uniform $\cdots$ could prevent bullying based on appearance or economic background.'& Claim: School uniform reduces bullying\\
        \midrule
        \textbf{Binary Question:} & {} \\
        Question 1: Would you think the snippet is relevant to the claim & Answer: YES\\
        Question 2:
        Would you think the snippet supports the claim & Answer: YES\\
        \bottomrule
        \end{tabular}
    }
    \caption{Annotate Template for Task 1}
    \label{annotation_task1}
\end{table*}

\section{Formula of Contrastive learning}
\label{sec:appendix_cl}
 Analogous to \citet{alshomary2021key}, we define the contrastive loss as follows:

\begin{align*}
    \mathcal {L}(A,P,N) =\  \max (&\ {\| f (A)-{f} (P)\|}_{2}\  -\\
    &\ {\| f (A)- {f} (N)\|}_{2}\ +\\
    &\ \alpha, 0\ )
\end{align*}

Here, $A$ represent the concatenation of topic and claim, $P$ is the positive input (i.e. convincing evidence), $N$ is the negative input (i.e. non-convincing evidence), $\alpha$ reflects the margin between positive and negative pairs, and $f$ is an embedding.

\section{Experiment Setting}
\label{sec:experiment_setting}
To establish benchmarks for our ASE datasets, we trained the baseline models from the previous section. All of the proposed models were trained and tested on two NVIDIA Tesla A100 80GB GPUs. All classification models were trained for 15 epochs (batch size of 64 (large), 32 (3B) and 8 (11B) respectively), while the summary model was trained for 5 epochs (batch size of 16 (large) and 4 (3B) respectively). The maximum input length for the summary task is 1024 and the max target length is 512. Other parameters are set by default.  All contrastive learning models were trained for 10 epochs with the training batch size as 32 for \texttt{Roberta-large} and \texttt{Setence-t5-large}, 8 for \texttt{Sentence-t5-3B}, respectively. The learning rate was set as $3e^{-5}$ and the maximum input sequence length was 256. We used Euclidean distance as the measurement between two embeddings in the contrastive loss function with $\alpha$ as 5 by default. The model with the best accuracy on the validation set was taken as the final trained model for inference.

\section{Project Debater's output}
\begin{figure*}
    \centering
    \includegraphics[scale = 0.2]{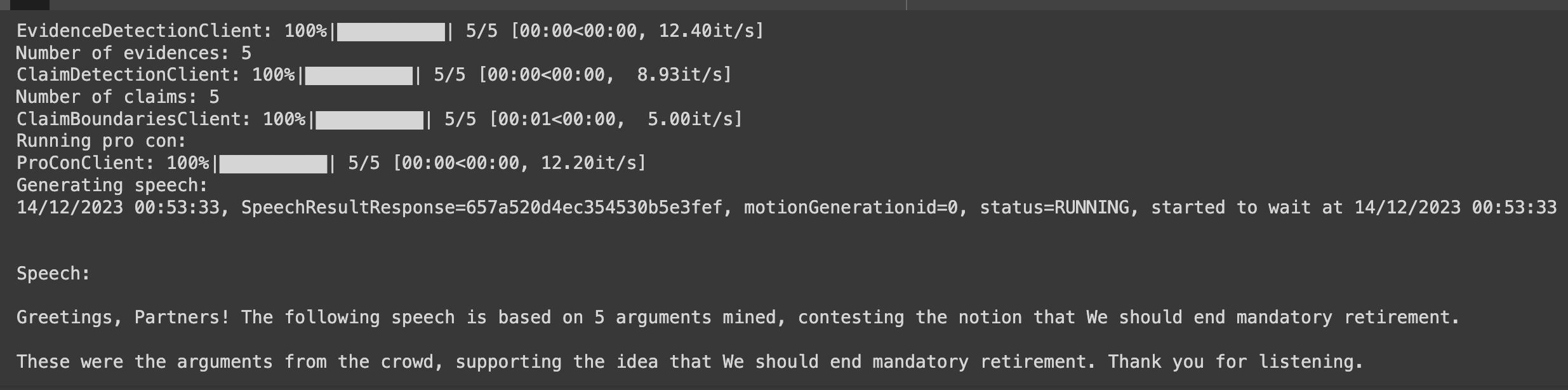}
    \caption{Output of narrative generation of Project Debater}
    \label{fig:enter-label}
\end{figure*}

Figure \ref{fig:enter-label} shows the result of IBM Project Debater.
\label{sec:project_output}

\section{Ablation Experiment}
\label{sec:Ablation Experiment}
\textit{Quality of Settings:} Table \ref{task3_irtresult} at Appendix \ref{sec:bt_irt} demonstrates the performance models as judged by humans and interpreted with IRT. 
Overall, the gold setting is significantly better than the evidence chosen by the baseline model (c.f. Average Golden vs Average Baseline), suggesting that there is scope for improvement on this challenging task. In the gold setting, the scenario that performs best occurs when all evidence is readily accessible  (c.f. average All, Best and Top2 scores). 
This suggests that the more comprehensive the model, the more effectively it can assess the quality of evidence, especially when the data is of high quality. In such cases, the model benefits from the additional information provided by the inputs. However, when supplied with irrelevant evidence, the models' performances tend to deteriorate due to the increased presence of noise (compare Top to All).

\textit{Performance of Models:} When it comes to specific models, as can be seen from the output of BT (From left to right: GPT4, GPT-3.5, GPT3, Bard, Alpaca, Vicuna = \{9.2483,9.1009,6.0515,-1.4266,-1.2101,-1.0872\}), 
GPT4 performed best in all settings 
followed by GPT-3.5 
while Alpaca 
and Vicuna 
performed worst, which demonstrates that more powerful models perform better in this task. Bard performs poorly, maybe due to the fact that it will disregard instructions to generate both supporting and opposing text where the source of the information cannot be identified, in the case of supporting evidence provided only.


\section{Example of whole dataset}

Table \ref{tab:example_alloftasks} shows the overview of whole dataset
\begin{table*}[t]
    \centering
    \scalebox{0.85}{
    \begin{tabular}{p{0.2cm}p{11.5cm}ccc}
    \toprule
    { } & \textbf{Topic: Shall we prohibit flag burning?}  & \textbf{Claim}  & \textbf{Evidence} & \textbf{Ranking}\\
    \midrule
    1 & Flags are important symbols that should be respected. & $C_1$ & {} & {} \\ 
    2 & Flag burning is an inflammatory act. & $C_2$ & {} & {}\\
    \hdashline
    3 & Warren was of the opinion that laws against flag-burning are constitutional. & {}  & $C_1$-$E_1$& Better \\
    4 & Proponents of legislation to proscribe flag burning argue that burning the flag is a very offensive gesture that deserves to be outlawed. & {}  & $C_2$-$E_1$ & Best\\
    5 & According to the ordinance in Hong Kong, a flag may not be used in advertisements or trademarks, and that publicly and wilfully burning, mutilating, scrawling on, defiling or trampling the flag is considered flag desecration. & {}  & $C_2$-$E_2$ & Worse \\
    6 & Rehnquist argued that flag burning is no essential part of any exposition of ideas but rather the equivalent of an inarticulate grunt or roar that, it seems fair to say, is most likely to be indulged in not to express any particular idea, but to antagonize others. & {}  & $C_2$-$E_3$ & Moderate \\
    \midrule
    {} & \textbf{Summary: Select the appropriate evidence(s) to draw up a debate script} & \textbf{Suff} & \textbf{Fact-Err} & \textbf{Ranking}\\
    \midrule
    7 & Supporters of the prohibition believe that flags are important symbols that must be respected. They point to the opinions of \colorbox{danshuhong}{justices like} Warren who considered laws against flag-burning to be constitutional. Further, they argue that flag burning is an inflammatory act, an offensive gesture that should be outlawed. They insist that to allow such an act would be to condone disrespect and provocation, potentially leading to societal discord. & Yes & Yes & {2} \\
    \bottomrule
    \end{tabular}
    }
    \caption{Examples of the full range of tasks included in the proposed dataset in a positive stance. The claims are derived from publicly known data that already exist, and the evidence is handpicked and paired with the claims by human experts (Task1). Six different models were asked to select the most convincing claim and evidence (Task2) to write a speech (Task3). All summaries were finally annotated by the human experts for argument sufficiency, the presence of factual errors (e.g. the word \colorbox{danshuhong}{justices like} in the example is hallucination), and the ranking of quality among all summaries.}
    \label{tab:example_alloftasks}
\end{table*}

\begin{table*}[htbp]
    \centering
    \scalebox{0.65}{
        \begin{tabular}{p{16cm}p{1cm}}
        \toprule
        Example Annotate Template For Task 3 & {} \\
        \midrule
        \textbf{Input Instance} & Ranking \\
        \midrule
        \textbf{Topic:} Routine child vaccinations should be mandatory & {} \\
        \midrule
        The topic of mandatory routine child vaccinations is a hotly contested one. Those opposing it bring forth several reasons, starting with the potential dangers and side effects associated with vaccines. For instance, the Rotashield vaccine was withdrawn from the market after links to higher instances of bowel obstructions were discovered, providing a chilling example of how vaccines can sometimes go wrong. Opponents also argue that mandating vaccinations would infringe upon the basic rights of individuals. They cite the First Amendment of the US Constitution, arguing that the mandate could interfere with the right to exercise personal and religious beliefs freely. The necessity of vaccinations is another point of contention. Critics point out that, according to the Mayo Clinic, natural infection often provides more robust immunity than a series of vaccinations, challenging the idea that routine vaccinations are essential to keep children healthy. & 1 \\
        \midrule
        Sure, here is a summary of the debate on whether routine child vaccinations should be mandatory: Pro-mandatory vaccination Vaccinations are safe and effective. They have been proven to prevent a variety of serious diseases, including measles, mumps, rubella, polio, and tetanus. Vaccinations are essential for public health. When a large percentage of the population is vaccinated, it creates herd immunity, which protects those who cannot be vaccinated, such as infants and people with weakened immune systems. Mandatory vaccination is a necessary way to protect public health. It is the only way to ensure that everyone who can be vaccinated is vaccinated, and that herd immunity is maintained. Anti-mandatory vaccination Vaccinations can have serious side effects, such as allergic reactions and neurological disorders. Mandatory vaccination violates individual rights. People have the right to make their own decisions about their health care, including whether or not to vaccinate their children. There are other ways to protect public health, such as education and personal hygiene. Conclusion The debate on whether routine child vaccinations should be mandatory is a complex one. There are valid arguments on both sides of the issue. Ultimately, it is up to each individual to decide whether or not to vaccinate their children. However, it is important to be informed about the risks and benefits of vaccination before making a decision. Here are some additional points that may be relevant to the debate: The risk of serious side effects from vaccination is very low. According to the Centers for Disease Control and Prevention (CDC), the risk of a serious side effect from a vaccine is about one in a million. The benefits of vaccination are much greater than the risks. Vaccinations have helped to save millions of lives and prevent countless cases of serious illness. Mandatory vaccination is not a new idea. It has been used for many years to protect public health. There are many ways to educate people about the benefits and risks of vaccination. This can be done through schools, health care providers, and other community organizations. It is important to have a balanced discussion about this issue. We should not demonize people who choose not to vaccinate their children. We should also not dismiss the concerns of those who are hesitant to vaccinate. By working together, we can find ways to protect public health while also respecting individual rights. & 2\\
        \midrule
        \#\#\# Assistant: The opponents of routine child vaccinations argue that mandatory vaccinations are dangerous. They point to the case of Rotashield, a vaccine for rotavirus, which was pulled from the market by the manufacturer after it was discovered that the vaccine might have contributed to higher instances of intussusception (bowel obstruction). They also argue that mandatory vaccination contradicts basic rights, citing the First Amendment of the US Constitution. Finally, they argue that routine child vaccinations are not necessary to keep children healthy, citing the fact that natural infection often provides more complete immunity than a series of vaccinations.
        \#\#\# Human: What kind of arguments can be made to counter the opponents to routine child vaccinations? 
        \#\#\# Assistant: There are several arguments that can be made to counter the opponents of routine child vaccinations: 1. Vaccines are safe and effective: While there may be rare side effects to vaccines, the benefits of vaccination far outweigh the risks. Vaccines have saved millions of lives and prevented countless illnesses and diseases. 2. Vaccines protect not just the individual but also the community: Vaccines are not just about protecting individual children, but also about protecting the entire community. By ensuring that children are vaccinated, we can reduce the spread of diseases and protect those who are too young to be vaccinated or those who cannot receive vaccines due to medical reasons. 3. Vaccines are essential for public health: Vaccines are an essential tool for preventing and controlling infectious diseases. Without vaccines, we would be at much greater risk of outbreaks and epidemics. 4. Mandatory vaccinations are necessary: In some cases, mandatory vaccinations may be necessary to protect public health. When there is a high risk of outbreaks or epidemics, mandatory vaccinations can help to prevent the spread of disease. 5. Vaccines are a personal choice: While some may argue that mandatory vaccinations infringe on their personal freedom, it is important to remember that vaccines are a personal choice that affects not just the individual but also the community. Ultimately, the decision to vaccinate should be based on the best available scientific evidence and the advice of healthcare professionals." & 3 \\

        \bottomrule
        \end{tabular}
    }
    \caption{Annotate Template for Task 3}
    \label{annotation_task3}
    \end{table*}

\end{document}